\documentclass[acmsmall,screen]{acmart}
    
\usepackage{graphicx}

\setcopyright{none}
\renewcommand\footnotetextcopyrightpermission[1]{}

\title{The Changing Role of Symbolic Methods in Artificial Intelligence}

\author{Jun Sun}
\affiliation{
  \institution{Singapore Management University}
  \city{Singapore}
  \country{Singapore}
}
\email{jun.sun@smu.edu.sg}

\begin{document}

\begin{abstract}

Why do intelligent systems need to reason? Computer science has long
regarded reasoning as a defining characteristic of intelligence, yet recent
advances in foundation models increasingly blur the distinction between
learning and reasoning. This article argues that the more fundamental
question is not \emph{how} intelligent systems reason, but \emph{why}
explicit symbolic reasoning is needed in the first place.

We propose the \emph{Compression Principle}: every computational model is a
simplified representation of reality, and explicit symbolic reasoning
compensates for information omitted during that simplification. From this
principle we derive the \emph{Modeling--Reasoning Trade-off}: as
computational models preserve richer representations of the world, the need
for explicit symbolic reasoning correspondingly decreases. This perspective
offers a unified explanation for both the historical success of symbolic
methods and the remarkable effectiveness of modern foundation models.

Paradoxically, the same development increases the importance of symbolic
methods for humans. As intelligent systems become more capable and less
transparent, symbolic representations increasingly serve as interfaces
through which humans specify requirements, verify behavior, regulate
autonomous systems, and establish trust. We therefore argue that the future
of symbolic methods lies not primarily as the computational engine of
machine intelligence, but as the symbolic interface between increasingly
capable AI systems and the humans who build, govern, and depend upon them.

\end{abstract}

\maketitle

\begin{quote}
\centering
\emph{``The map is not the territory.''}\\
---Alfred Korzybski
\end{quote}

%%%%%%%%%%%%%%%%%%%%%%%%%%%%%%%%%%%%%%%%%%%%%%%%%%%%%%%%%%

\section{Why Do Intelligent Systems Reason?}

Symbolic reasoning has long been regarded as one of the defining characteristics of
intelligence. From Aristotle's syllogisms to modern formal verification,
from theorem proving to automated planning, computer science has devoted
decades to developing increasingly sophisticated methods for symbolic
reasoning. Indeed, much of our discipline is built upon the premise that
intelligent behavior arises from the ability to derive new knowledge from
existing knowledge through logical inference
\cite{newell1976,mccarthy1960,russell2021}.

Recent advances in artificial intelligence (AI), however, challenge this
long-standing perspective. Foundation models routinely generate software,
translate natural languages, answer scientific questions, and increasingly
interact with the physical world---often without relying on explicitly
constructed symbolic knowledge bases or handcrafted reasoning rules
\cite{vaswani2017,brown2020,openai2023}. Their remarkable capabilities have
reignited a familiar debate. Should future AI systems continue to rely on
symbolic reasoning? Will statistical learning ultimately replace symbolic
methods? Or will intelligence emerge from some combination of the two?

These questions have inspired decades of research. Yet they all share an
implicit assumption: that reasoning itself is a fundamental component of
intelligence. This article begins by questioning that assumption.

Rather than asking \emph{how} intelligent systems reason, we ask a more
fundamental question:

\begin{quote}
\centering
\emph{Why do intelligent systems need to perform explicit symbolic reasoning
at all?}
\end{quote}

The distinction is subtle but important. Questions about \emph{how} lead us
to design better reasoning algorithms, richer logical systems, or more
expressive symbolic representations. Questions about \emph{why} instead ask
what computational role explicit reasoning actually serves. If that role can
be explained more fundamentally, then the changing role of symbolic methods
in modern AI may also become easier to understand.

Throughout this article, we return repeatedly to a simple thought
experiment. Imagine approaching a busy road intersection. A human driver determines whether to proceed primarily by observing a
traffic light. The symbolic signal---red, yellow, or green---reduces an
enormously complicated traffic situation to a decision that can be made
almost instantly.

Now imagine the same intersection occupied entirely by autonomous vehicles.
Each vehicle continuously observes its surroundings, exchanges information
with nearby vehicles, predicts future trajectories, and constructs a rich
computational model of the entire traffic environment. Does such a system
still require traffic lights? Or can every vehicle determine directly
whether it is safe to proceed?

The question is not merely about transportation. It illustrates a much
broader issue. Why do symbolic abstractions exist in the first place? We argue that they arise because intelligent systems have historically been
forced to operate on simplified models of reality. Explicit symbolic
reasoning compensates for information omitted during that simplification.
From this observation we derive a simple principle explaining why symbolic
reasoning became central to computer science, why increasingly rich AI
models reduce its computational role, and why symbolic methods are
nevertheless becoming more important than ever for human interaction with
AI.

The future of symbolic methods, we argue, is therefore neither dominance
nor disappearance. Instead, their role is fundamentally changing—from the
computational engine of intelligent systems to the symbolic interface
between increasingly capable AI systems and the humans who build, govern,
and depend upon them.

%%%%%%%%%%%%%%%%%%%%%%%%%%%%%%%%%%%%%%%%%%%%%%%%%%%%%%%%%%

\section{The Compression Principle}

The traffic-light example illustrates a simple but surprisingly general
observation. Before making a decision, an intelligent system must first
decide \emph{what information to represent}. Only after a representation has
been constructed can computation begin.

This distinction is fundamental. Computer science has traditionally focused
on algorithms, inference procedures, and reasoning mechanisms. Yet every
algorithm operates on a computational model rather than on reality itself.
Whether the model is a logical theory, a finite-state machine, a software
specification, a knowledge graph, or a neural network, it necessarily
captures only selected aspects of the world while ignoring others.

This observation leads to what we call the \emph{Compression Principle}.

\begin{quote}
\textbf{Compression Principle}

\emph{Every computational model is a simplified representation of reality.
Explicit symbolic reasoning compensates for information omitted during model
construction.}
\end{quote}

The principle separates intelligent computation into two distinct stages. The first stage is \emph{model construction}. Information from the external
world is transformed into a computational representation. During this
process, countless details are discarded because they are irrelevant,
redundant, unavailable, or simply too expensive to retain. The second stage is \emph{reasoning}. Given the resulting computational
model, new conclusions are derived that are not explicitly represented.
Reasoning therefore reconstructs information that is useful for decision
making but absent from the simplified representation.

\begin{figure*}[t]
    \centering
    \includegraphics[width=\textwidth]{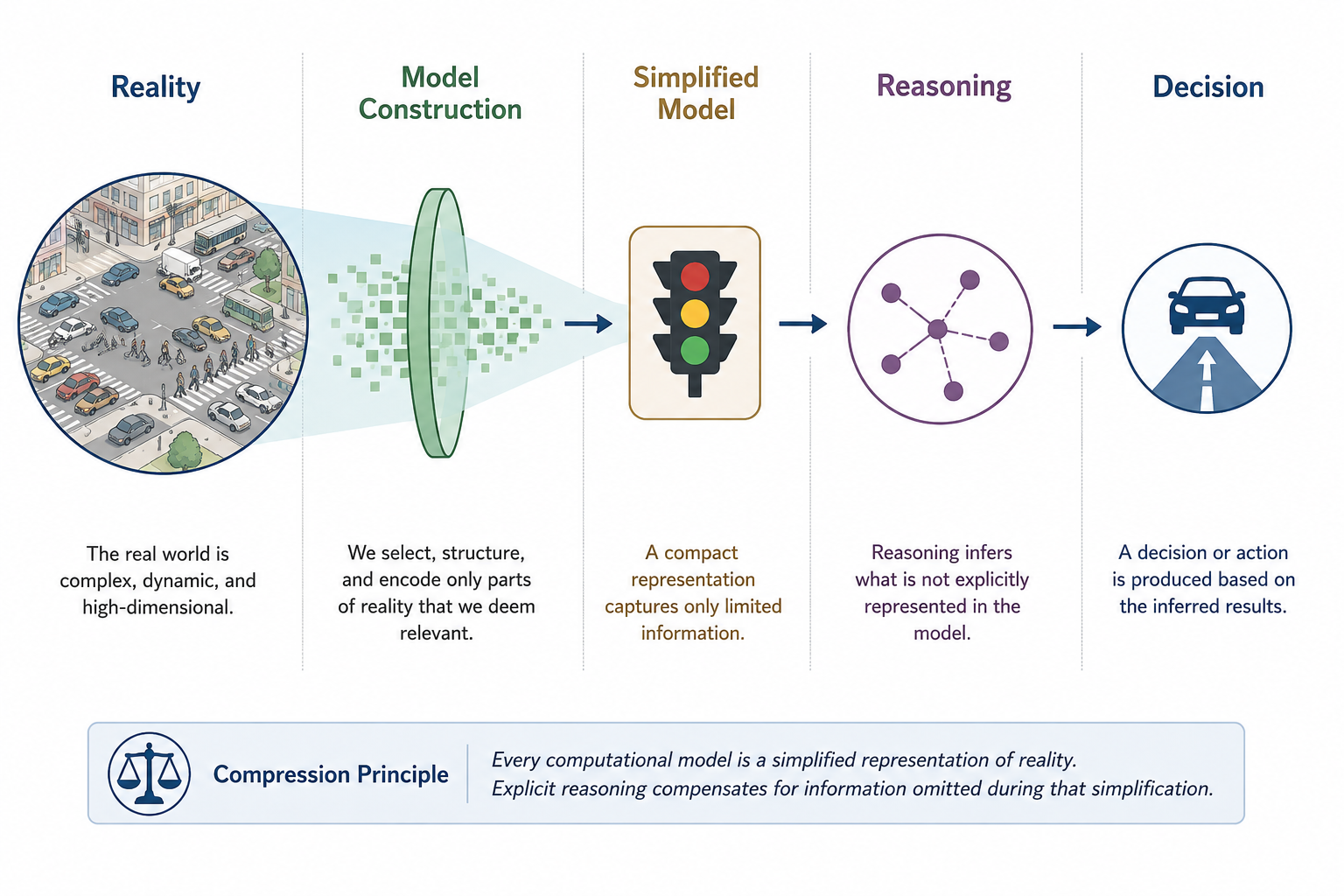}
    \caption{The \emph{Compression Principle}.}
    \label{fig:compression}
\end{figure*}

Figure~\ref{fig:compression} summarizes this perspective. Returning to the traffic intersection makes this distinction concrete.
The complete traffic environment contains enormous amounts of information:
vehicle positions, velocities, accelerations, road conditions, pedestrian
movements, visibility, weather, and driver intentions. Human drivers cannot
process this information directly. Instead, the traffic control system
compresses the environment into one of three symbolic states: red, yellow,
or green.

Notice what has happened. The complexity has not disappeared. Rather, it has been hidden inside the
model construction process. The driver subsequently performs explicit
reasoning over an extremely small symbolic representation. The reasoning is
simple precisely because the representation has already removed most of the
complexity.

This example illustrates both the strengths and the limitations of symbolic
methods. Their strength derives directly from simplification. Simpler models require
less memory, are easier to communicate, admit efficient reasoning
algorithms, and often expose elegant mathematical structure. Much of
computer science has advanced by discovering abstractions that preserve the
information relevant to a computational task while discarding everything
else. Programming languages abstract away hardware details. Software
specifications suppress implementation complexity. Finite-state machines
replace continuous dynamics with discrete states. Scientific theories model
only the variables believed to matter.

Their limitations arise from exactly the same source. Whenever different real-world situations become represented by the same
symbolic abstraction, that distinction is permanently lost. No amount of
reasoning can reconstruct information that was never represented in the
first place. Logical inference may therefore be perfectly correct with
respect to the computational model while nevertheless producing conclusions
that fail in reality. George Box's famous observation that ``all models are
wrong, but some are useful'' \cite{box1978} is therefore not merely a
statement about models. It is also a statement about the limits of reasoning
performed on those models.

The Compression Principle also provides a different perspective on the
long-standing debate between symbolic AI and statistical learning.
Traditionally, these approaches have been viewed as fundamentally different
computational paradigms. We suggest instead that they differ primarily in
where computational effort is invested.

Classical symbolic systems devote substantial effort to explicit reasoning
over highly simplified representations. Modern foundation models devote
substantial effort to constructing extraordinarily rich computational
representations through large-scale learning. In both cases, intelligence
emerges from computation performed on a model. What changes is how much
information is preserved before explicit reasoning begins.

This observation immediately suggests a broader engineering principle. As computational models preserve increasingly rich representations of the
world, the amount of explicit symbolic reasoning required after model
construction should correspondingly decrease. We refer to this relationship as the \emph{Modeling--Reasoning Trade-off},
which forms the basis for the remainder of this article.

%%%%%%%%%%%%%%%%%%%%%%%%%%%%%%%%%%%%%%%%%%%%%%%%%%%%%%%%%%

\section{Bounded Computation}

The Compression Principle and the Modeling--Reasoning Trade-off naturally
raise an important question. If richer computational models reduce the need
for explicit symbolic reasoning, why has symbolic reasoning occupied such a
central position throughout the history of computer science?

The answer lies in bounded computation. Throughout most of the history of computing, computational resources have
been scarce. Memory was limited, processors were slow, sensors captured only
small amounts of information, and collecting large datasets was often
impractical. Under these constraints, representing the full richness of the
real world was simply impossible. Intelligent systems therefore had little
choice but to construct highly simplified computational models and rely on
explicit reasoning to compensate for the information that those models
necessarily omitted.

The same observation applies to human intelligence. Humans possess limited memory, limited attention, and limited reasoning
capacity. Herbert Simon described this phenomenon as \emph{bounded
rationality}: intelligent agents make decisions under computational
constraints rather than under conditions of perfect knowledge
\cite{simon1957,simon1979}. From the perspective of the Compression
Principle, bounded rationality can be understood as a special case of
bounded computation. Humans simplify the world because they cannot process
its full complexity, and explicit symbolic reasoning enables them to operate
effectively on these simplified mental models.

Returning to the traffic-light example illustrates this point. A human driver approaching an intersection cannot continuously estimate the
future trajectories of every nearby vehicle, predict pedestrian behavior,
evaluate road conditions, and determine whether crossing is safe. The
traffic-light system therefore performs a substantial amount of computation
on the driver's behalf. It compresses a highly complex traffic environment
into one of three symbolic states. The driver's reasoning task is greatly
simplified because much of the complexity has already been absorbed into the
symbolic abstraction.

This pattern appears throughout computer science. Programming languages replace machine instructions with high-level
abstractions. Software specifications replace implementation details with
behavioral requirements. Finite-state machines replace continuous dynamics
with discrete states. Scientific theories replace enormously complicated
physical phenomena with compact mathematical descriptions. Database schemas
replace the complexity of the real world with entities, attributes, and
relationships.

These abstractions differ in purpose and representation, yet they all follow
the same computational strategy. They deliberately simplify reality in order
to make computation feasible. Explicit symbolic reasoning subsequently
operates on the simplified representation rather than on reality itself.

Importantly, this should not be interpreted as a weakness of symbolic
methods. Quite the opposite. Symbolic abstractions have been extraordinarily
successful because they provide remarkably efficient compromises between
expressiveness and computational tractability. Much of the progress of
computer science has depended on discovering abstractions that discard
precisely the information that can safely be ignored while preserving the

Seen from this perspective, the historical dominance of symbolic methods
becomes much less surprising. They did not become central because symbolic
reasoning is the only path to intelligence. Rather, they became central
because bounded computation made aggressive simplification unavoidable.

This observation immediately suggests a new question. If the computational limitations that historically required aggressive
simplification continue to diminish, how should we expect the role of
explicit symbolic reasoning to evolve? The next section argues that this
transition is already taking place.

%%%%%%%%%%%%%%%%%%%%%%%%%%%%%%%%%%%%%%%%%%%%%%%%%%%%%%%%%%

\section{The Age of Rich Models}

The Compression Principle and the Modeling--Reasoning Trade-off suggest a
simple prediction. As computational resources increase, intelligent systems
should be able to construct increasingly rich computational models of the
world. As these models preserve more information, the need for explicit
symbolic reasoning after model construction should correspondingly decrease.

This prediction describes a broad historical trend in computer science.
For decades, progress was often achieved by designing increasingly
sophisticated reasoning algorithms over carefully engineered symbolic
representations. Expert systems encoded domain knowledge as rules.
Automated planners reasoned over symbolic states and actions. Computer
vision relied on handcrafted features and geometric models. Machine
translation depended on manually designed grammars and linguistic rules.
The central challenge was to compensate, through explicit reasoning, for
information that could not feasibly be represented.

Increasing computational resources have fundamentally altered this balance.
Modern computing systems routinely process enormous datasets, store vast
amounts of information, and construct computational models containing
billions or even trillions of parameters. Rather than investing
computational effort primarily in designing increasingly sophisticated
reasoning procedures, modern AI increasingly invests computation in model
construction itself. Rich representations learned from data replace many of
the manually engineered abstractions that earlier systems depended upon
\cite{krizhevsky2012,hinton2012,vaswani2017,brown2020,openai2023}.

Foundation models represent perhaps the clearest example of this broader
transition. They construct rich statistical representations of language,
source code, images, and other modalities directly from massive amounts of
data. Many tasks that historically required explicit symbolic reasoning can
now be solved by operating directly on these learned representations.
Reasoning has not disappeared; rather, much of the information that
previously had to be reconstructed through explicit symbolic inference is
already captured by the computational model itself.

Returning once more to the traffic-light example illustrates this
transition. Imagine an intersection populated entirely by autonomous vehicles. Each
vehicle continuously observes the environment, exchanges information with
other vehicles, predicts future trajectories, and maintains an accurate
computational model of the entire intersection. The decision problem no
longer begins with a symbolic traffic signal. Instead, it begins with a
rich model of the surrounding world. Consequently, the question changes. Instead of asking
\begin{center}
\emph{``Is the traffic light green?''}
\end{center}
the vehicle asks
\begin{center}
\emph{``Given my current model of the environment, is it safe to
proceed?''}
\end{center}
The symbolic abstraction has become unnecessary because the computational
model itself preserves sufficient information for making the decision.

This example illustrates the Modeling--Reasoning Trade-off introduced in
the previous section. As model fidelity increases, explicit symbolic
reasoning becomes responsible for progressively less of the overall
computation. The computational burden shifts from deriving missing
information to constructing increasingly accurate models in the first
place.

Importantly, this does not imply that reasoning disappears. Every
computational model remains an abstraction of reality and therefore omits
some information. Rich models continue to require inference, planning,
prediction, and decision making. What changes is the role of \emph{explicit
symbolic reasoning}. Instead of compensating for substantial information
loss, it increasingly fills only the relatively small gaps that remain
after rich model construction.

This perspective also sheds new light on the long-standing debate between
symbolic AI and statistical learning. Rather than viewing the two as
competing paradigms, we suggest viewing them as different allocations of
computational effort. Classical symbolic systems devote substantial effort
to reasoning over simplified models. Modern AI systems devote substantially
more effort to constructing rich computational models, thereby reducing the
need for explicit symbolic reasoning afterwards.

Seen from this perspective, the recent success of foundation models is not
an isolated technological breakthrough, but the latest stage of a broader
evolution in computing. As computational resources continue to grow,
constructing richer models increasingly becomes a more effective investment
than designing increasingly elaborate reasoning procedures over simplified
ones.

This conclusion, however, leads directly to an apparent paradox. If
explicit symbolic reasoning is becoming less central to intelligent
computation, why are symbolic methods becoming more prominent in
discussions of AI safety, explainability, governance, and regulation? The
answer lies not in the computational needs of machines, but in the
cognitive needs of humans.

%%%%%%%%%%%%%%%%%%%%%%%%%%%%%%%%%%%%%%%%%%%%%%%%%%%%%%%%%%

\section{From Engine to Interface}

The previous section argued that increasingly rich computational models
reduce the need for explicit symbolic reasoning within intelligent systems.
Viewed purely from the perspective of machine computation, one might
therefore conclude that symbolic methods are gradually becoming obsolete.

We argue that it might instead be the opposite. The apparent contradiction arises because the Compression Principle applies
not only to machines, but also to humans. Although modern AI systems are rapidly becoming capable of constructing rich
computational models of the world, human cognitive capabilities remain
fundamentally bounded. Our memory, attention, and reasoning capacity have
changed little despite dramatic advances in computing. Humans continue to
understand complex systems by constructing simplified symbolic models of
their behavior.

Consequently, the role of symbolic methods is changing. Historically, symbolic methods primarily served the computational needs of
machines. Programming languages enabled computation. Logical theories
supported automated inference. Planning languages represented actions and
goals. Expert systems encoded domain knowledge as symbolic rules.

Increasingly, however, symbolic methods serve a different purpose. They provide the abstractions through which humans understand, supervise,
verify, regulate, and ultimately trust increasingly capable AI systems.

The traffic-light example illustrates this transition. Consider again an intersection populated entirely by autonomous vehicles.
Each vehicle maintains a detailed computational model of the surrounding
traffic environment and continuously coordinates with nearby vehicles. From
the perspective of the vehicles, symbolic traffic lights may no longer be
necessary. Every vehicle can determine directly whether it is safe to
proceed.

Now consider the same intersection from a human perspective. A pedestrian approaching the crossing cannot inspect the internal
representations of hundreds of interacting AI systems. A traffic engineer
cannot directly verify millions of continuously changing trajectories. A
city regulator cannot legislate over high-dimensional neural
representations. Humans therefore continue to require symbolic abstractions
through which they understand and coordinate the behavior of the system.

The traffic light has not disappeared. Its purpose has changed. It is no longer primarily part of the vehicles' computational process.
Instead, it has become an interface between autonomous machines and the
humans who live alongside them.

We believe the same transition is occurring throughout AI. Increasingly capable AI systems may rely internally on rich learned
representations with relatively little explicit symbolic reasoning.
Nevertheless, humans continue to demand symbolic specifications, formal
contracts, safety properties, explanations, regulations, and ethical
principles. These symbolic artifacts need not mirror the internal
computation performed by the AI system. Rather, they express human
expectations in forms that people can inspect, debate, verify, revise, and
enforce.

Viewed in this way, many active research areas acquire a different
interpretation. Formal verification is not merely about helping machines reason correctly;
it provides mathematical guarantees that humans can understand and trust.
Explainable AI is not necessarily about exposing every internal computation
performed by a neural model; it is about constructing explanations that are
meaningful to human users. AI governance is not primarily concerned with how
machines internally compute, but with how human societies specify
acceptable behavior and assign responsibility.

These observations suggest that symbolic methods are undergoing a
fundamental migration. They are moving away from the computational core of intelligent systems
toward the boundary between machine intelligence and human society.

In other words,
\begin{quote}
\centering
\textbf{Symbols are not intelligence. Symbols are interfaces.}
\end{quote}
This statement should not be interpreted literally. Intelligent systems will
continue to employ symbolic representations whenever they provide
computational advantages. Rather, it captures a broader shift in emphasis.
The historical importance of symbolic methods lay primarily in enabling
intelligent computation under bounded computational resources. Their future
importance may lie increasingly in enabling effective communication between
bounded humans and increasingly capable intelligent machines.

This perspective resolves the apparent paradox developed throughout this
article. The computational importance of explicit symbolic reasoning may continue to
decline even as the societal importance of symbolic methods continues to
grow. Their future therefore lies neither in replacing modern AI nor in
disappearing from it, but in becoming the symbolic interface through which
humans understand, govern, and collaborate with intelligent systems.

%%%%%%%%%%%%%%%%%%%%%%%%%%%%%%%%%%%%%%%%%%%%%%%%%%%%%%%%%%

\section{Implications for Computer Science}

The perspective developed in this article suggests that the long-standing
debate between symbolic AI and statistical learning may be framed too
narrowly. Rather than asking whether future intelligent systems should be
symbolic or neural, we should ask a more fundamental question:

\begin{quote}
\centering
\emph{Where should symbolic methods reside in increasingly capable AI
systems?}
\end{quote}

For much of the history of computing, the answer was straightforward.
Symbolic methods formed the computational core of intelligent systems. They
represented knowledge, performed reasoning, generated plans, and derived
decisions. This historical role reflected the computational constraints of
the time. Simplified representations were unavoidable, making explicit
symbolic reasoning indispensable.

As computational models become richer, however, the center of gravity begins
to shift. Symbolic methods increasingly migrate away from the internal
computational mechanisms of AI systems toward their interfaces with humans.
This shift has important implications for several areas of computer science.

First, it suggests a changing role for formal methods. Specifications,
contracts, invariants, runtime monitors, and verification need not describe
the internal computations performed by an AI system. Instead, they define
the behavioral properties that humans expect intelligent systems to satisfy.
Formal methods therefore become increasingly concerned with specifying and
verifying externally observable behavior rather than reconstructing internal
reasoning processes.

Second, it suggests a complementary view of explainable AI. Much current
research attempts to recover the internal reasoning of large AI models. Our
perspective suggests that explanations need not faithfully reproduce every
internal computation. Their primary purpose is communication. A useful
explanation is one that enables humans to understand, predict, and trust a
system's behavior, even if it abstracts away much of the underlying
computation.

Third, it suggests that AI safety and governance should be viewed primarily
as problems of symbolic interface design. Regulations, policies, safety
requirements, ethical principles, and legal obligations are all symbolic
artifacts through which societies communicate expectations to intelligent
systems. As AI becomes increasingly autonomous, designing interfaces between
human intentions and machine behavior may prove just as important as
improving the intelligence of the underlying models themselves.

More broadly, the Compression Principle suggests a useful way to reconcile
two traditions that have often been viewed as competing. Rich learned models
and symbolic methods solve different computational problems. Rich models
preserve information required for intelligent behavior. Symbolic methods
enable bounded humans to understand, supervise, coordinate, and regulate
that behavior.

This perspective therefore points toward a different research agenda.
Rather than attempting to replace one paradigm with another, future AI
research may increasingly focus on the interaction between the two:
constructing richer computational models while simultaneously developing
more expressive symbolic interfaces through which humans can communicate
with, reason about, and govern increasingly capable intelligent systems.

If this perspective is correct, the future of symbolic methods is not one of
decline, but of migration. Their greatest contributions may increasingly
occur not inside intelligent systems, but at the boundary between machine
intelligence and human society.

%%%%%%%%%%%%%%%%%%%%%%%%%%%%%%%%%%%%%%%%%%%%%%%%%%%%%%%%%%

\section{Conclusion}

We began this article with a simple question:

\begin{quote}
\centering
\emph{Why do intelligent systems need to perform explicit symbolic reasoning?}
\end{quote}

We have argued that explicit symbolic reasoning is not a defining property
of intelligence, but a computational consequence of operating on simplified
models of reality. The \emph{Compression Principle} explains why symbolic
reasoning has historically been so successful, while the
\emph{Modeling--Reasoning Trade-off} explains why increasingly rich
computational models reduce its computational role.

This perspective also resolves an apparent paradox. As AI systems become
more capable, the need for explicit symbolic reasoning within those systems
may continue to diminish. Yet symbolic methods themselves become
increasingly important because humans remain computationally bounded.
Specifications, explanations, regulations, and formal guarantees provide
the symbolic interfaces through which humans understand, supervise, and
govern increasingly capable intelligent systems.

Returning one final time to the traffic-light example illustrates this
transformation. Future autonomous vehicles may no longer require traffic
lights to coordinate safely, because each vehicle possesses a sufficiently
rich model of its surroundings. Traffic lights may nevertheless remain—not
because machines need them, but because humans do.

We therefore suggest a different way of viewing the future of symbolic
methods. Their role is evolving from the computational engine of
intelligent systems to the symbolic interface between machine intelligence
and human society. The future of AI is therefore unlikely to be symbolic or
neural. Rather, it will increasingly combine rich computational models with
symbolic interfaces that enable humans to understand, verify, and shape the
behavior of intelligent machines.

%%%%%%%%%%%%%%%%%%%%%%%%%%%%%%%%%%%%%%%%%%%%%%%%%%%%%%%%%%

\section*{Acknowledgment}

The author used AI-based assistants as part of the research and writing
process to brainstorm ideas, explore alternative explanations, challenge
assumptions, and improve the clarity of the manuscript. All substantive
ideas, arguments, interpretations, and conclusions presented in this
article represent the author's own views. Any errors, omissions, or
shortcomings are solely the responsibility of the author.

\bibliographystyle{ACM-Reference-Format}
\bibliography{references}

\end{document}